\documentclass[9pt, conference]{IEEEtran}
\IEEEoverridecommandlockouts
\usepackage{cite}
\usepackage{amsmath,amssymb,amsfonts}
\usepackage{algorithmic}
\usepackage{graphicx}
\usepackage{textcomp}
\usepackage{xcolor}

\usepackage{multirow}
\usepackage{booktabs}
\usepackage[ruled,linesnumbered]{algorithm2e}
\usepackage[normalem]{ulem}
\usepackage{balance}

\def\BibTeX{{\rm B\kern-.05em{\sc i\kern-.025em b}\kern-.08em
    T\kern-.1667em\lower.7ex\hbox{E}\kern-.125emX}}
\begin{document}

\title{AppealNet: An Efficient and Highly-Accurate Edge/Cloud Collaborative Architecture for DNN Inference
}

\author{\IEEEauthorblockN{Min Li\IEEEauthorrefmark{1}, Yu Li\IEEEauthorrefmark{1}, Ye Tian\IEEEauthorrefmark{1}, Li Jiang\IEEEauthorrefmark{2} and Qiang Xu\IEEEauthorrefmark{1}}
\IEEEauthorblockA{\IEEEauthorrefmark{1}\textit{\underline{CU}hk \underline{RE}liable Computing Laboratory (CURE)}, \\
\textit{Department of Computer Science and Engineering},
\textit{The Chinese University of Hong Kong},
Shatin, Hong Kong S.A.R.\\
\{mli, yuli, tianye, qxu\}@cse.cuhk.edu.hk}
\IEEEauthorblockA{\IEEEauthorrefmark{2}\textit{Department of Computer Science and Engineering, MoE Key Lab of Artificial Intelligence, AI Institute}, \\ 
\textit{Shanghai Jiao Tong University}, Shanghai, China 
}
}

\maketitle

\begin{abstract}
This paper presents \emph{AppealNet}, a novel edge/cloud collaborative architecture that runs deep learning (DL) tasks more efficiently than state-of-the-art solutions. For a given input, AppealNet \emph{accurately} predicts on-the-fly whether it can be successfully processed by the DL model deployed on the resource-constrained edge device, and if not, appeals to the more powerful DL model deployed at the cloud. This is achieved by employing a two-head neural network architecture that explicitly takes inference difficulty into consideration and optimizes the tradeoff between accuracy and computation/communication cost of the edge/cloud collaborative architecture. 
Experimental results on several image classification datasets show up to more than 40\% energy savings compared to existing techniques without sacrificing accuracy.
\end{abstract}


\section{Introduction}
\label{sec:intro}



Over the years, internet-of-things (IoTs) have become increasingly intelligent by employing deep neural network (DNN) models for learning-based processing. Due to the stringent resource constraints of such edge devices, the DNN models used in them must be small.  
To address this problem, various DNN model compression techniques (\textit{e.g.}, \textit{weight pruning}~\cite{he2017channel}, \textit{quantization}~\cite{han2015deep}, \textit{knowledge distillation}~\cite{hinton2015distilling}) 
and compact DNN architectures~\cite{howard2017mobilenets} were proposed for edge computing in the literature. 

Small DNN models are usually less accurate than large ones due to their limited model capacities. Generally speaking, small models have difficulty handling those corner cases that lie in the long-tail of data distribution, but they can still infer correctly for the majority of the other inputs. Consequently, it is possible to employ an edge/cloud collaborative architecture that processes those `easy' inputs with a small DNN model at edge and offloads `difficult' inputs to the powerful DNN model at cloud~\cite{tann2016runtime, park2015big, bolukbasi2017adaptive, stamoulis2018design}. Such an architecture facilitates to achieve a good energy-accuracy tradeoff for DNN inference when compared to edge-only or cloud-only solutions. 

\begin{figure}[t]
	\centering
	\includegraphics[width=\linewidth]{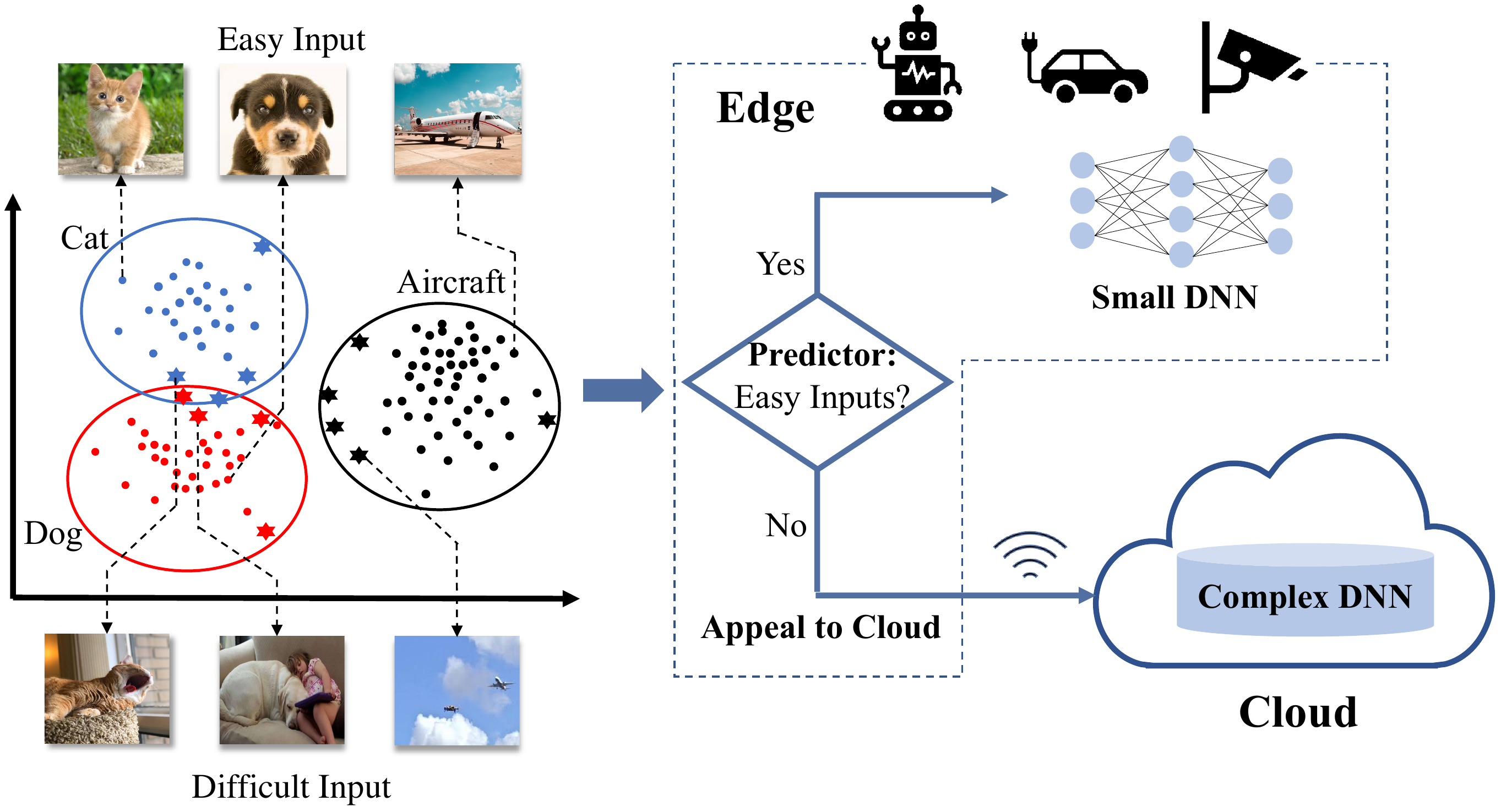}
	\setlength{\abovecaptionskip}{-5pt}
	\caption{Edge/Cloud Collaborative Architecture}
	\label{fig:bl-flow}
	\vspace{-5pt}
\end{figure}

No doubt to say, the effectiveness of such an edge/cloud collaborative architecture depends on the accuracy of the \emph{predictor} that is responsible for differentiating `easy' and `difficult' inputs for the edge DNN model (see Fig.~\ref{fig:bl-flow}). Mis-predicted `easy' inputs result in accuracy loss while mis-predicted `difficult' inputs cause unnecessary communication and computation cost.
Existing works typically rely on the ``confidence" level of the model as the main indicator and design the predictor accordingly~\cite{tann2016runtime, park2015big, bolukbasi2017adaptive, stamoulis2018design}. 
However, such kind of softmax probability-based confidence measurement has been shown to be inaccurate, as DNNs tend to produce overconfident incorrect predictions~\cite{hendrycks2016baseline, guo2017calibration, lakshminarayanan2017simple}.


In this work, we propose to explicitly model the inference difficulty (instead of implicitly estimate a `confidence' level) when training the small edge neural network. To be specific, considering the small edge DNN as an approximator for the large cloud DNN, we develop a two-head architecture for the small edge DNN model that outputs the approximation results and the prediction results simultaneously, namely \emph{AppealNet}.
The edge/cloud collaborative architecture exploration problem is then formulated as a joint optimization problem to optimize the tradeoff between accuracy and cost, wherein we co-optimize the approximator design and the predictor design on when to appeal to the complex DNN model. We prove that this optimization problem can be solved with a novel training loss representation form, enabling trustworthy differentiation between `easy' and `difficult' inputs. To the best of our knowledge, this is the first work that provides theoretical foundation for edge/cloud collaborative architecture design.

To demonstrate the effectiveness of AppealNet, we perform extensive experiments on several image classification tasks using various DNN models. The experimental results show that AppealNet is efficient and highly accurate for DNN inference, achieving up to more than 40\% energy savings than existing methods, without any accuracy loss. 

We organize the remainder of this paper as follows. We review related works in Section~\ref{sec:related}.
Section \ref{sec:bl} introduces the problem of edge/cloud collaboration. In Section~\ref{sec:objective}, we present the optimization objective of AppealNet. Next, we detail the AppealNet architecture and the training method in Section~\ref{sec:appealnet}. Then, Section~\ref{sec:exp} presents our experimental results. Finally, Section~\ref{sec:conc} concludes this paper.

\section{Related Work}
\label{sec:related}


Applications such as autonomous driving, video surveillance, and speech recognition generate zillions of bytes of data at the network edge, and it is essential to conduct learning to achieve edge intelligence. As edge devices are usually resource-limited, there are both \textit{static} and \textit{dynamic} techniques towards effective deployments of DNNs onto them.

Static techniques include various types of model compression techniques (\textit{e.g.}, weight pruning~\cite{he2017channel}, quantization~\cite{han2015deep}, knowledge distillation~\cite{hinton2015distilling}) and compact model designs~\cite{howard2017mobilenets}.
Given an edge device, they adjust the hyper-parameters of the small edge DNN model (\textit{e.g.}, compression ratio, data types, and channel width) to fit into the device at minimum accuracy degradation from an accurate large model. Dynamic techniques treat incoming data differently and perform selective computations considering that a large portion of inputs do not require a complex DNN model to process.

We can further categorize dynamic DNN solutions into two categories. The first category performs selective computations with model partitioning. For example,~\cite{wang2018skipnet, wang2020dual, panda2016conditional, long2020conditionally} selectively skip some of the computational layers and blocks by inserting extra decision gates in the model. 
BranchyNet~\cite{teerapittayanon2016branchynet}, dynamic routing~\cite{liu2018dynamic}, and shallow-deep network~\cite{kaya2018shallow} early-exit the computation when certain classification confidence from layer-wise outputs is achieved.
Moreover, adjusting the network depth according to the input scale provides an alternative way for dynamic processing~\cite{yang2020resolution}. While partitioning the complex DNN model across the edge device and cloud does reduce computational cost, it may incur significant communication overheads as the intermediate outputs of complex DNNs could be several orders of magnitude larger than the original raw input~\cite{he2016deep}. 

The second category employs multiple DNN models in the system and dynamically selects an appropriate one for processing each input at runtime~\cite{park2015big, bolukbasi2017adaptive, stamoulis2018design,nan2017adaptive}.
Park \emph{et. al}~\cite{park2015big} proposes a Big/Little neural network design for edge/cloud collaboration. They use the the difference between the largest and the second-largest softmax confidence value as the indicator to choose which model to use. Later, \cite{bolukbasi2017adaptive, stamoulis2018design,nan2017adaptive} extend it to a series of models and optimize the collaborative architecture for energy savings. 

The Big/Little architecture proposed in~\cite{park2015big} is attractive, but using softmax  probability-based confidence for model selection is questionable. Several works (\textit{e.g.},~\cite{hendrycks2016baseline, guo2017calibration}) have shown that the predictive softmax-based distribution from DNNs has a poor relationship with the actual inference confidence. For example, high-confidence prediction are frequently produced for many mis-classified inputs. Therefore, it is essential to develop new indicators to differentiate those `easy' inputs and `difficult' inputs for edge/cloud collaboration. 
\section{Edge/Cloud Collaboration Problem}
\label{sec:bl}




Let us consider a robot vacuum cleaner that is equipped with an image classifier for obstacle avoidance as the edge device. Let input image $\mathbf{x} \in \mathcal{X}$ be the data collected from the cleaner's camera, and its corresponding classification label $y \in \mathcal{Y}$ (\textit{e.g.}, pets, chairs, or tables) be a random variable that follows a joint data distribution $P(\mathbf{x}, y)$ (the joint distribution of images and the corresponding labels from a specific house), in which $\mathcal{Y}=\{1,\cdots, K\}$,  $K$ refer to the number of classes, and $y$ is the ground-truth label for input $\mathbf{x}$. 

Since the inputs are collected in a dynamic and non-ideal condition, the objects in the images do not always behave in the predictable way. For instance, cats may show up in strange poses (e.g. yawning in Fig.~\ref{fig:bl-flow}), which are beyond the prediction capability of the small model equipped locally in the robot. Given this, if we allow the edge devices to make all predictions locally, the accuracy might be not satisfactory, and further robot actions based on the wrong predictions may have some negative impact. To achieve higher accuracy, an edge/cloud collaborative architecture can be used. 

To form such a collaborative architecture, we assume two discriminant models (\textit{e.g.}, convolutional neural networks) are trained on a dataset 
drawn \textit{i.i.d} from $P(\mathbf{x}, y)$. One is a computationally-intensive neural network using lots of resources at Cloud to achieve high performance, denoted by $f_0: \mathcal{X}\to \mathcal{Y}$. The other is a small edge neural network (\textit{i.e.}, an \textit{approximator} of the large model), with limited capacity and lower accuracy, denoted by $f_1: \mathcal{X}\to \mathcal{Y}$. 
During the inference stage, a \textit{soft} prediction function (\textit{i.e.}, \textit{predictor}) $q(z|\mathbf{x})\in [0, 1]$ with $z \in \{0,1\}$, which is also deployed on the robot, is used to decide whether the small DNN $f_1$ is adequate to classify the input $\mathbf{x}$ (\textit{e.g.}, common objects in the house without much variation), or the input should be off-loaded to the powerful Cloud DNN model for processing $f_0$. To be specific, we adopt the \textit{threshold-based} approach, and assign label $1$ to $z$ if $q(z|\mathbf{x})$ is above some threshold $\delta$ (\textit{e.g.}, $0.5$), and label it $0$, otherwise. We expect the predictor to be sufficiently accurate so that only those `difficult' inputs are off-loaded to the Cloud.

Thus, an edge/cloud collaborative architecture is a combination of ($f_0, f_1, q$), and the final output of edge/cloud collaborative architecture with respect to a specific input $\mathbf{x}$ is:

\begin{equation}
	(f_0, f_1, q)(\mathbf{x}) = 
		\begin{cases}
		f_1(\mathbf{x})          & \text{if } q(1|\mathbf{x}) \ge \delta, \\
		f_0(\mathbf{x})		     & \text{otherwise.}
		\end{cases}
	\label{eq:bl}
\end{equation}
\quad The performance of a stand-alone model $f_z$ could be modeled by $\ell (f_z(\mathbf{x}), y), z \in \{0, 1\}$, where $\ell: \mathcal{X} \times \mathcal{Y} \to \mathbb{R}^+$ is a task loss function, for example, \textit{cross-entropy} loss which is often used in classification tasks. 
Therefore, taking the respective loss 
of the complex and small DNN into consideration, the overall expected loss of the edge/cloud collaborative architecture can be calculated by Eq.~\eqref{eq:risk}.
\begin{multline}
	\qquad \qquad \qquad 
	\mathbb{E}_{P(\mathbf{x},y)} \mathbb{E}_{q(z|\mathbf{x})}[\ell (f_z(\mathbf{x}),y)] = \\
	\mathbb{E}_{P(\mathbf{x},y)}[q(1|\mathbf{x})\cdot \ell (f_1(\mathbf{x}),y)
	+ 
	(1-q(1|\mathbf{x}))\cdot \ell (f_0(\mathbf{x}),y)]
	\label{eq:risk}
\end{multline}
\begin{multline}
	\qquad \qquad \qquad
	\mathbb{E}_{P(\mathbf{x},y)} \mathbb{E}_{q(z|\mathbf{x})}[cost (f_z, q, \mathbf{x})] =\\
	\mathbb{E}_{P(\mathbf{x},y)}[q(1|\mathbf{x})\cdot cost (f_1, q, \mathbf{x}) 
	+ (1-q(1|\mathbf{x}))\cdot cost (f_0, q, \mathbf{x})]
	\label{eq:cost}
\end{multline}
\quad Eq.~\eqref{eq:cost} shows the cost of the edge/cloud collaboration system. For a given input $\mathbf{x}$, the cost of using a particular model for prediction is denoted by $cost(f_z, q, \mathbf{x})$ (to be detailed in Section~\ref{subsec:or})

In most edge applications, especially for resource-limited IoT devices, a budget constraint must be imposed on the system. In the edge/cloud collaboration setting, thus, an optimal design can be defined as minimizing the overall expected loss, under a particular cost constraint $b$. We formally write the optimization objective as follows:
\begin{multline}
	\qquad \qquad
	\min_{f_0,f_1 \in \mathcal{F}, q \in \mathbf{Q}}
	\mathbb{E}_{P(\mathbf{x},y)} \mathbb{E}_{q(z|\mathbf{x})}[\ell (f_z(\mathbf{x}),y)] \\
	s.t. \quad
	\mathbb{E}_{P(\mathbf{x},y)} \mathbb{E}_{q(z|\mathbf{x})}[cost (f_z, q, \mathbf{x})] \le b \qquad
	\label{eq:original}
\end{multline}

Note that, the key difference between our formulation and previous work~\cite{park2015big, bolukbasi2017adaptive, stamoulis2018design} is that, the predictor function is parameterized and can be co-optimized when training the edge/cloud collaborative models.


\section{Proposed Relaxation Method}
\label{sec:objective}

In this section, we relax the optimization objective of edge/cloud collaboration to make it tractable to solve. 
We first consider the case that the complex network is a pre-trained white-box model, and separate it from the optimization objective. Then, we look into the cost constraints of ($f_0, f_1, q$) and simplify them as a single constraint with respect to the expectation of $q(z|\mathbf{x})$.
Next, we consider the black-box scenario where the complex DNN is a black box during optimization and present how to relax the optimization objective in this case. Note that, without loss of generality, we consider the optimization objective of classification tasks with cross-entropy loss in the following discussion.

\subsection{Objective Relaxation}
\label{subsec:or}
As shown in Eq.~\eqref{eq:original}, our objective is to minimize the expected overall loss of the edge/cloud collaborative DNN models, under a given constraint on the total system cost. The output of the optimization procedure is the combination of ($f_0, f_1, g$). 

The main challenge for solving this optimization problem is its huge solution space. In practice, the complex DNN could be an ensemble of various models or a very deep network to achieve state-of-the-art performance. Consequently, optimizing $f_0$ and $f_1$ together may lead to very slow convergence. 
To reduce complexity, we assume a known fixed complex DNN $f_0$ with high accuracy. Note that, the fixed complex DNN is available when training the small edge DNN so that we could get the loss item $\ell (f_0(\mathbf{x}), y)$ for optimization. 

\vspace{5pt}
\textbf{Cost constraints.} 
As shown in Eq.~\eqref{eq:cost}, with the predictor function $q$, the cost to infer an input $\mathbf{x}$ depends on which DNN to execute it. For a static DNN, the cost for its inference is independent of the specific input $x$. Therefore, we have $cost(f_z, q, \mathbf{x})=cost(f_z, q)$. 

Let us consider the standalone DNN model first. 
The main cost include off-chip memory access of DNN parameters and neuron computation. As both are directly related to the computational demands, 
we can simply use computational cost (in unit of FLOPS) to approximate them. 

With edge/cloud collaboration, some inputs are off-loaded to the cloud and the results are then sent back to the edge device. Thus, running the predictor at the edge and the communication between the edge and the cloud also contributes to the total cost. We can use a constant $c_1 = cost(f_1, q)$ to represent the cost of running predictor/small DNN on the edge device and another constant $c_0=cost(f_0, q)$ to represent the accumulated cost of running predictor on the edge device, running complex DNN on the cloud, and the communication overhead.

Often the related cost of processing the input by the complex DNN on the cloud is much larger than that by the small edge DNN. Therefore, we have use $c_0 \gg c_1$. By substituting Eq.~\eqref{eq:cost} with $c_0$ and $c_1$, we have the reduced cost function as Eq.~\eqref{eq:cost1}. 
\begin{equation}
	(c_1 - c_0) \cdot \mathbb{E}_{P(\mathbf{x},y)}[ q(1|\mathbf{x})] + c_0
	\label{eq:cost1}
\end{equation}
As the cost is under a specific budget $b$ ($c_1<b<c_0$), we can derive the simplified constraint as Eq.~\eqref{eq:constraint2}. 
\begin{equation}
\mathbb{E}_{P(\mathbf{x},y)} [q(1|\mathbf{x})] \ge \hat{b}, \hat{b} = \frac{c_0 - b}{c_0 - c_1}
\label{eq:constraint2}
\end{equation}
\quad According to Eq.~\eqref{eq:constraint2}, $q(1|\textbf{x})$ is an indicator of the overall cost, and the new cost constraint $\hat{b} \in(0, 1)$ ensures at least $\hat{b}$ fraction of inputs would be processed by the small edge DNN.

\vspace{5pt}
\textbf{Relaxed optimization objective.}
With the above, we have the constrained optimization objective of edge/cloud collaborative design as follows:
\begin{multline}
	\qquad \quad
	\min_{f_1 \in \mathcal{F}, q \in \mathbf{Q}}
	\mathbb{E}_{P(\mathbf{x},y)} \mathbb{E}_{q(z|\mathbf{x})}[\ell (f_z(\mathbf{x}),y)] \\
	s.t. \quad
	\mathbb{E}_{P(\mathbf{x},y)} [q(1|\mathbf{x})] \ge \hat{b} \qquad \qquad \qquad
	\label{eq:final}
\end{multline}
To convert the constraint into objective itself, we further take $log()$ of predictor likelihood $q(z|\mathbf{x})$, so that the constraint becomes:
\begin{equation}
log(\hat{b}) - \mathbb{E}_{P(\mathbf{x},y)}[log(q(1|\mathbf{x}))] \le 0
\label{eq:final-ct}
\end{equation}
\quad Re-writing Eq.~\eqref{eq:final} as a \textit{Lagrangian} under \textit{KKT} conditions~\cite{kuhn2014nonlinear}, we can put the constraint~\eqref{eq:final-ct} into the objective function, remove the constant term $log(\hat{b})$, and obtain the final optimization objective as follows:
\begin{equation}
	\min_{f_1, q}
	\mathbb{E}_{P(\mathbf{x},y)} \mathbb{E}_{q(z|\mathbf{x})}[\ell (f_z(\mathbf{x}),y)] + 
	\beta \cdot \mathbb{E}_{P(\mathbf{x},y)}[-log(q(1|\textbf{x}))])
	\label{eq:final-re}
\end{equation}
\subsection{Black-Box Approximation}
\label{subsec:bba}
In the earlier discussion, we assume the complex DNN is a white-box neural network, where its output information is available so that we could calculate the classification loss $\ell (f_0(\textbf{x}), y)$ for a  specific input.  However, the complex DNN might be unknown during optimization, in many cases. For example, in our edge-cloud collaboration scenarios, the complex DNN might be provided by a machine learning service vendor running at a remote data center, and it is agnostic to the small DNN to be deployed on the edge. 



Under such circumstances, we propose to treat the black-box complex DNN as an oracle function $\mathcal{O}(\cdot)$, \textit{i.e.}, it can correctly classify every input from $P(\mathbf{x},y)$. 
Now, the loss term $\ell (f_0(\mathbf{x}),y)=\ell (\mathcal{O}(\mathbf{x}),y)$ in Eq.~\eqref{eq:final-re} becomes zero as the outputs of the oracle function are always the ground-truth values. Therefore, the optimization objective becomes:
\begin{equation}
	\min_{f_1, q}
	\mathbb{E}_{P(\mathbf{x},y)}[q(1|\mathbf{x}) \cdot \ell (f_1(\mathbf{x}),y)] + 
	\beta \cdot \mathbb{E}_{P(\mathbf{x},y)}[-log(q(1|\textbf{x}))]
	\label{eq:final-re-m}
\end{equation}

\section{AppealNet Design}
\label{sec:appealnet}
In this section, we present the overall architecture of AppealNet and its training method, in which we jointly train the little network and the predictor in an end-to-end manner. We also illustrate the full workflow for AppeaNet.

\subsection{Two-Head Architecture}
\begin{figure}[t]
	\centering
	\includegraphics[width=0.5\linewidth]{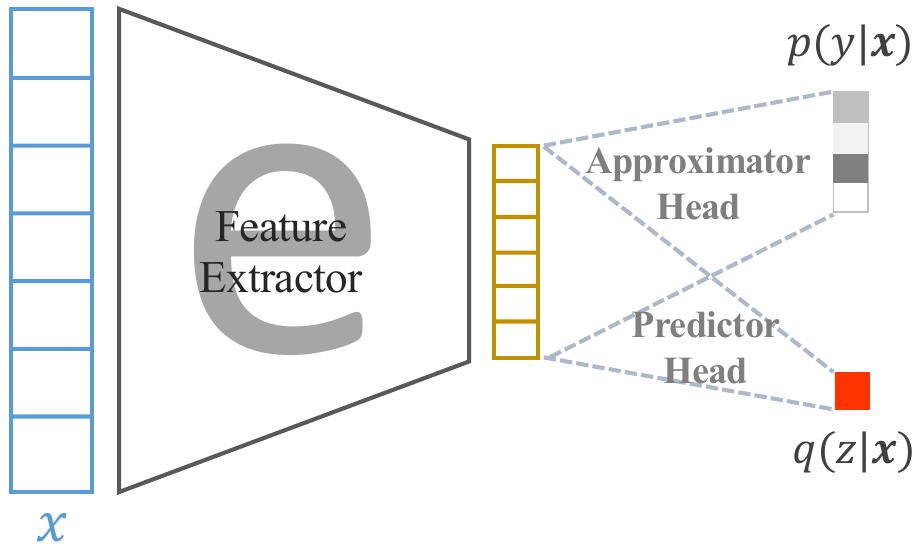}
	\caption{Two-head architecture for the little network in AppealNet.}
	\label{fig:cb-archi}
	\vspace{-15pt}
\end{figure}

Fig.~\ref{fig:cb-archi} demonstrates the two-head design architecture, which consists of one common feature extractor and two separate heads (\textit{i.e.}, \textit{approximator head}  and \textit{predictor head}). Using a two-head design has several benefits. First, merging $\mathcal{F}$ and $\mathcal{Q}$ into one unified design space enables us to utilize and  modify \textit{off-the-shelf} efficient DNN architectures (\textit{e.g.}, EfficientNet~\cite{TanL19}, ShuffleNet~\cite{ZhangZLS18}, MobileNet~\cite{howard2017mobilenets}) for the approximator/predictor design. For a given device with specific resource constraints, we can first select an efficient DNN architecture that meets this constraint and add a predictor head to the original architecture with minimal overhead.
Second, such a design facilitates training $(f_0, q)$ jointly in an end-to-end manner. 

To be more specific, we parameterize the predictor function $q(z|\mathbf{x})$ as a neural network, and enforce $f_1$ and $q$ sharing all computational-intensive layers (\textit{e.g.}, convolutional layers) wrapped in the \textit{feature extractor}. Thus, the approximator consists of the feature extractor and an approximator head, while the predictor consists of the feature extractor and a predictor head. For classification tasks, the approximator head contains several cascaded fully-connected layers and a \textit{softmax} layer to output the predictive probability vector $p(y|\mathbf{x})$, wherein 
the $i^{\text{th}}$ element indicates the probability of the $i^{\text{th}}$ class. For the predictor head used in AppealNet, we simply employ a single fully-connected layer to keep the design overhead minimum. The output of predictor passes through a \textit{sigmoid} activation function such that $q(z|\mathbf{x}) \in [0, 1]$.

\subsection{Joint Training Method}
\label{subsec:jtm}


Alg.~\ref{algo:jt} shows the joint training scheme, which optimizes the approximator $f_1$ and the predictor $q$ jointly with the help of the big network $f_0$. To be specific, we initialize the network parameters  using the pre-trained little network without the predictor head first. This step facilitates us to take advantage of available model zoos (\textit{e.g}, PyTorch model zoo) and accelerate the training process (line 1). Then we insert the predictor head into the original network and start training (line 2). 
For each batch of training samples, we calculate the loss $\mathcal{L}$, which consists of the system loss and system cost (see line 5). The loss is back-propagated to update the parameters of $(f_1, q)$. These training steps are repeated until convergence is reached. 

\begin{algorithm}[t]
	\caption{Joint Training for Two-Head Small Network.}
	\label{algo:jt}
	\KwIn{ Dataset $\textbf{S}$ sampled from $P(\textbf{x},y)$, 
		Big Network $f_0$}
	\KwOut{Little Network $(f_1, q)$.}
	\tcc{Initialize}
	 Initialize $f_1$ with the pre-trained model\;
	 Insert the predictor head and get two-head little network $(f_1, q)$\;
	\tcc{Define Loss Functions; $\ell$ depends on tasks and settings}
	$\mathcal{L}_p = q(1|\mathbf{x})\cdot \ell (f_1(\mathbf{x}),y)+ (1-q(1|\mathbf{x}))\cdot \ell (f_0(\mathbf{x}),y)$ \;
	$\mathcal{L}_q = -log(q(1|\textbf{x}))$\;
	\tcc{Training Process}
	\While{Not Convergence}{
		Sample ${(\textbf{x}_1, y_1),\cdots,(\textbf{x}_M, y_M)}$ from $\textbf{S}$\;
		$\mathcal{L} = \frac{1}{M}\sum_{i}[\mathcal{L}_p(\textbf{x}_i, y_i|f_0,f_1,q)+
		\beta \mathcal{L}_q(\textbf{x}_i|q)]$\;
		Update $(f_1, q)$ by descending stochastic gradient of $\mathcal{L}$ \;
	}
	
	\Return Little Network$(f_1, q)$;
\end{algorithm}

\subsection{Full Workflow for AppealNet}

\begin{figure}[h]
	\centering
	\includegraphics[width=0.9\linewidth]{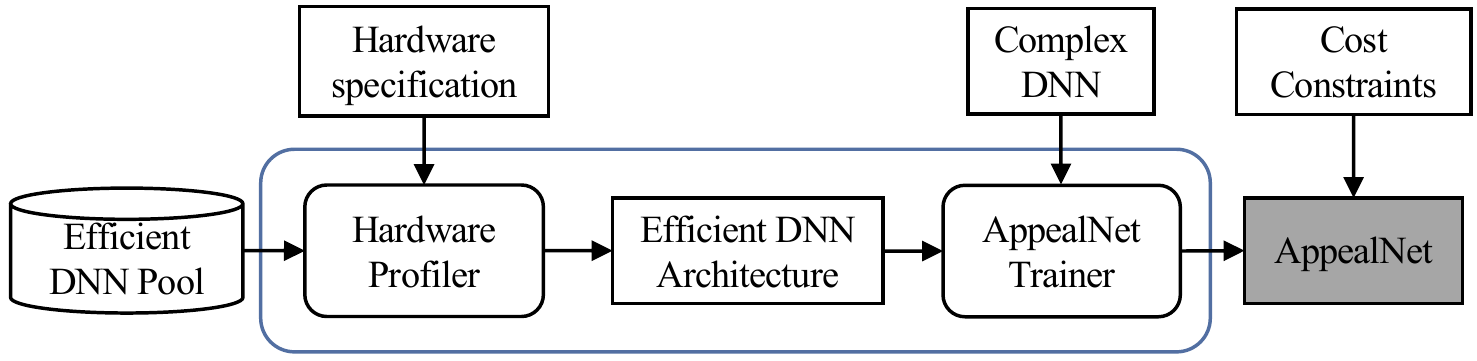}
	\caption{Working flow of AppealNet}
	\label{fig:workflow}
	\vspace{-5pt}
\end{figure}

Fig.~\ref{fig:workflow} shows an overview of the overall workflow of AppealNet. First, given the hardware specification and a pool of efficient DNN candidate models, the hardware profiler analyzes which DNN model meets the resource constraints of the targeted hardware platform, and outputs a suitable small edge DNN architecture. Next, the architecture is augmented by the predictor head and fed into the AppealNet trainer for joint training with the help of the complex DNN model, under a given cost constraint with edge/cloud collaboration.

\section{Experimental Results}
\label{sec:exp}

\begin{figure}[t]
	\vspace{-5pt}
	\centering
	\includegraphics[width=0.95\linewidth]{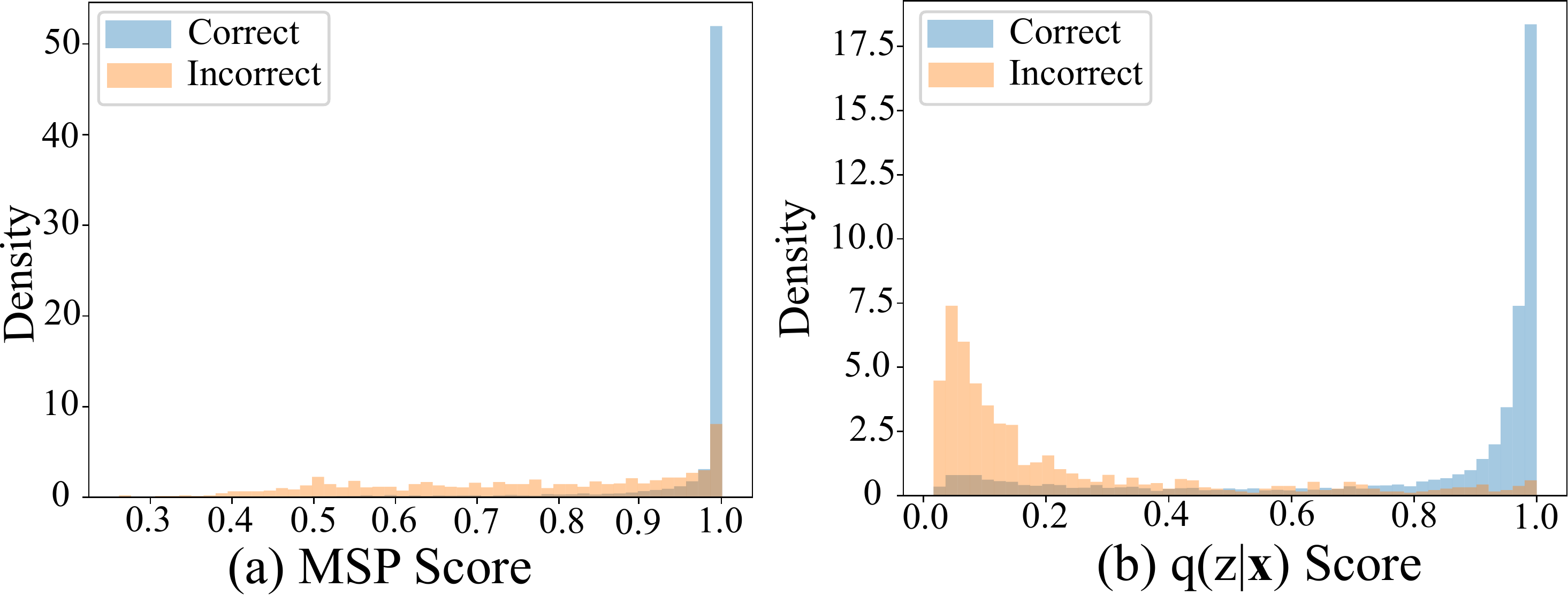}
	\caption{Histogram of MSP and AppealNet on CIFAR-10.}
	\vspace{-10pt}
	\label{fig:ce-msp}
\end{figure}

\subsection{Experimental Setup}
\textbf{Datesets:}
We demonstrate the effectiveness of AppealNet using four classification datasets: The German Traffic Sign Recognition Benchmark (GTSRB)~\cite{stallkamp2012man}, CIFAR-10, CIFAR-100~\cite{krizhevsky2009learning}, and Tiny-ImageNet~\cite{deng2009imagenet}. 


\textbf{Neural Networks: }
We evaluate AppealNet design by applying it to three off-the-shelf efficient CNN families: MobileNet~\cite{howard2017mobilenets}, EfficientNet~\cite{TanL19}, and ShuffleNet~\cite{ZhangZLS18}. They can be deployed on a diverse range of mobile/IoT devices with different hardware resource capacity. We choose ResNet-101~\cite{he2016deep} as the complex network.  

\textbf{Baseline Solutions: }
We consider conventional confidence-score based methods as baseline solutions. These baseline solutions mainly adopt the following score metrics to decide which neural network, the small edge DNN or the complex cloud DNN, to execute:
\begin{enumerate}
	\item Maximum softmax probability~(\textit{MSP})~\cite{hendrycks2016baseline} is defined as the maximum probability of a misclassified inputs. \textit{MSP} is obtained from the softmax function: $MSP = 1^{st}~p(y|\mathbf{x})$. 
	\item The gap/score margin~\cite{park2015big, bolukbasi2017adaptive, stamoulis2018design}~(\textit{SM}) is defined as the difference between the largest output probability and the second largest one: $SM = 1^{st}~p(y|\mathbf{x}) - 2^{nd}~p(y|\mathbf{x})$.
	\item The entropy score~\cite{teerapittayanon2016branchynet} (\textit{Entropy}) is derived by computing the entropy of the output predictive distribution of each input and further taking its minus: 
	$Entropy = \sum_{j}p(y|\mathbf{x})_j log[p(y|\mathbf{x})_j]$.
\end{enumerate}

\textbf{Evaluation Metrics: }
We define the skipping rate~($SR$) -- the fraction of inputs being selected by the predictor to process with the little network -- to evaluate the proposed predictor ($q(z|\mathbf{x})$):
\begin{equation}
\text{SR}= \frac{1}{N}\sum_{i}^N \mathbb{I}(q(1|\mathbf{x}_i)\ge \delta)
\label{eq:sr}
\end{equation}
\quad In contrast, the appealing rate ($AR$) represents the portion of inputs being off-loaded to the big network:
\begin{equation}
\text{AR}= \frac{1}{N}\sum_{i}^N \mathbb{I}(q(1|\mathbf{x}_i) < \delta)
\label{eq:ar}
\end{equation}
\quad The overall accuracy of the edge/cloud architecture for DNN inference can be calculated as follows:
\begin{multline}
\text{Acc.}(f_0, f_1, g) = \frac{1}{N} \sum_{i}^N [\mathbb{I}(q(1|\mathbf{x}_i)\ge \delta)\mathbb{I}(f_1(\textbf{x}_i)=y_i) \\+ \mathbb{I}(q(1|\mathbf{x}_i) < \delta)\mathbb{I}(f_0(\textbf{x}_i)=y_i)]
\label{eq:overall-acc}
\end{multline}
\quad We define the relative accuracy improvement ($AccI$) as the improvement of classification accuracy of the edge/cloud collaborative architecture compared with the stand-alone small DNN deployed on the edges, scaled by the accuracy gap between two DNNs.
\begin{equation}
\text{AccI} = \frac{\text{Acc.}(f_0, f_1, q) - \text{Acc.}(f_1)}{\text{Acc.}(f_0)-\text{Acc.}(f_1)}
\label{eq:acc-improvement}
\end{equation}
\quad We use the computational cost of the overall edge/cloud system for cost comparison: 
\begin{equation}
Cost_{overall} = SR \times cost(f_1, q) + (1 - SR) \times cost(f_0, q)
\label{eq:overall-cost}
\end{equation}
\quad Consider cloud-edge collaborative architectures, the bandwidth and latency cost of off-loading the data from edge to cloud are directly related to the skipping rate of inputs being processed by the small network on the edge; while the performance and energy cost of the whole system are evaluated by the overall accuracy improvement and cost listed above.


\begin{figure*}[t]
	\centering
	\includegraphics[width=\linewidth]{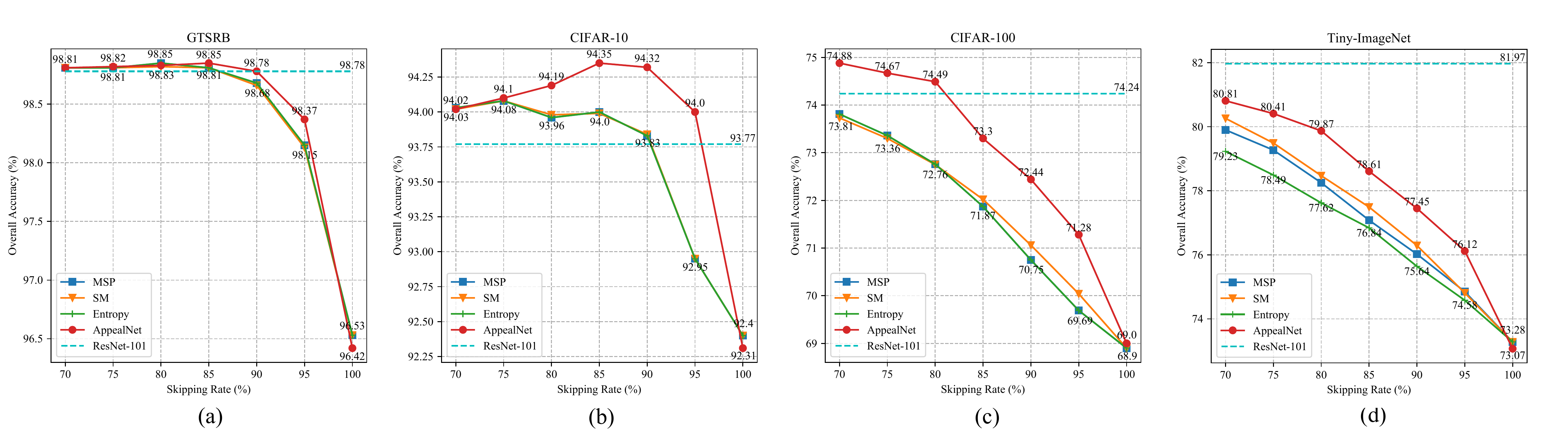}
	\caption{Overall accuracy under different skipping rate for MobileNet on GTSRB, CIFAR10, CIFAR100 and Tiny-ImageNet.}
	\label{fig:acc-four}
	\vspace{-5pt}
\end{figure*}

\begin{table*}[t]
	\resizebox{\linewidth}{!}{
\begin{tabular}{ccc|cccccccc}
\hline
\multirow{2}{*}{Dataset} & \begin{tabular}[c]{@{}c@{}}Accuracy\\ \%\end{tabular} & \begin{tabular}[c]{@{}c@{}}Cost\\ MFLOPs\end{tabular} & \begin{tabular}[c]{@{}c@{}}Cost\\ at 50.0\% AccI\end{tabular} & \multirow{2}{*}{\begin{tabular}[c]{@{}c@{}}Relative Saving\\ at 50.0\% AccI\end{tabular}} & \begin{tabular}[c]{@{}c@{}}Cost\\ at 75.0\% AccI\end{tabular} & \multirow{2}{*}{\begin{tabular}[c]{@{}c@{}}Relative Saving\\ at 75.0\% AccI\end{tabular}} & \begin{tabular}[c]{@{}c@{}}Cost\\ at 90.0\% AccI\end{tabular} & \multirow{2}{*}{\begin{tabular}[c]{@{}c@{}}Relative Saving\\ at 90.0\% AccI\end{tabular}} & \begin{tabular}[c]{@{}c@{}}Cost\\ at 95.0\% AccI\end{tabular} & \multirow{2}{*}{\begin{tabular}[c]{@{}c@{}}Relative Saving\\ at 95.0\% AccI\end{tabular}} \\ \cline{2-3}
                         & \multicolumn{2}{c|}{ResNet-101/MobileNet/AppealNet}                                                           & (SM/AppealNet)                                                &                                                                                           & (SM/AppealNet)                                                &                                                                                           & (SM/AppealNet)                                                &                                                                                           & (SM/AppealNet)                                                &                                                                                           \\ \hline
GTSRB                    & 98.78/96.53/96.42                                     & 2520.3/28.8/28.8                                      & 122.23/100.80                                                 & 17.53\%                                                                                   & 192.24/123.48                                                 & 35.77\%                                                                                   & 228.37/166.58                                                 & 27.07\%                                                                                   & 227.20/191.99                                                 & 15.50\%                                                                                   \\ \hline
CIFAR-10                 & 93.77/92.40/92.31                                     & 2520.3/94.6/94.6                                      & 217.10/138.02                                                 & 36.43\%                                                                                   & 245.72/161.31                                                 & 34.35\%                                                                                   & 291.32/164.70                                                 & 43.46\%                                                                                   & 314.13/167.86                                                 & 46.56\%                                                                                   \\ \hline
CIFAR-100                & 74.24/68.90/69.00                                     & 2520.4/94.7/94.7                                      & 415.38/234.66                                                 & 43.51\%                                                                                   & 634.418/385.54                                                & 39.22\%                                                                                   & 819.50/462.19                                                 & 43.60\%                                                                                   & 884.51/513.89                                                 & 41.90\%                                                                                   \\ \hline
Tiny-ImageNet            & \multicolumn{1}{l}{81.97/73.28/73.07}                 & \multicolumn{1}{l|}{7832.34/313.11/313.11}            & 1482.35/1067.29                                               & 28.00\%                                                                                   & 2215.48/1724.47                                               & 22.16\%                                                                                   & 3089.21/2646.33                                               & 14.34\%                                                                                   & 4374.25/3835.71                                               & 12.31\%                                                                                   \\ \hline
\end{tabular}
	}
	\vspace{5pt}
	\caption{Overall computational cost of edge/cloud architecture under different accuracy requirements.}
	\label{tb:baseline}
	\vspace{-10pt}
\end{table*}

\subsection{Visualization of Input Differentiation}
\label{subsec:vo}
In Figure~\ref{fig:ce-msp}, we visualize the MSP score and $q(z|\textbf{x})$ score with EfficientNet as the small edge network and the training dataset is CIFAR-10. The \textit{correct} and \textit{incorrect} inputs refer to those inputs that can be correctly processed at edge and those that cannot be processed, respectively. These two figures clearly show the advantage of AppealNet. That is, AppealNet is able to separate the "easy" inputs and "difficult" inputs well with respect to $q(z|\textbf{x})$ scores, while MSP scores\footnote{The results of SM scores and Entropy scores are similar to that of MSP scores.} lead to much more overlap between these two kinds of inputs. Such advantage is obtained with the customized predictor head in the AppealNet architecture and the corresponding training strategy with theoretical foundations. 
Similar phenomena are also observed for other datasets and models. 


\begin{table*}[t]
	
	\resizebox{\linewidth}{!}{
	\begin{tabular}{ccc|cccccccc}
\hline
Model        & \begin{tabular}[c]{@{}c@{}}Original \\ Accuracy\\ (\%)\end{tabular} & \begin{tabular}[c]{@{}c@{}}AppealNet \\ Accuracy\\ (\%)\end{tabular} & \begin{tabular}[c]{@{}c@{}}AR \\ at 50.0\% AccI\\ (SM/AppealNet)\end{tabular} & \begin{tabular}[c]{@{}c@{}}Relative Saving\\ at 50.0\% AccI\end{tabular} & \begin{tabular}[c]{@{}c@{}}AR \\ at 75.0\% AccI\\ (SM/AppealNet)\end{tabular} & \begin{tabular}[c]{@{}c@{}}Relative Saving\\ at 75.0\% AccI\end{tabular} & \begin{tabular}[c]{@{}c@{}}AR \\ at 90.0\% AccI\\ (SM/AppealNet)\end{tabular} & \begin{tabular}[c]{@{}c@{}}Relative Saving\\ at 90.0\% AccI\end{tabular} & \begin{tabular}[c]{@{}c@{}}AR \\ at 95.0\% AccI\\ (SM/AppealNet)\end{tabular} & \begin{tabular}[c]{@{}c@{}}Relative Saving\\ at 95.0\% AccI\end{tabular} \\ \hline
EfficientNet & 90.44                                                               & 91.95                                                                & 13.78/7.65                                                                    & 44.48\%                                                                  & 25.30/18.63                                                                   & 26.36\%                                                                  & 23.24/18.94                                                                   & 18.50\%                                                                  & 36.53/44.36                                                                   & 17.65\%                                                                  \\ \hline
MobileNet    & 92.40                                                               & 92.31                                                                & 6.22/4.18                                                                     & 32.80\%                                                                  & 12.49/16.74                                                                   & 25.39\%                                                                  & 18.94/24.51                                                                   & 22.72\%                                                                  & 25.59/34.59                                                                   & 26.02\%                                                                  \\ \hline
ShuffleNet   & 87.30                                                               & 87.35                                                                & 21.29/16.43                                                                   & 22.82\%                                                                  & 34.95/28.27                                                                   & 19.11\%                                                                  & 45.30/36.92                                                                   & 18.50\%                                                                  & 45.59/54.37                                                                   & 16.15\%                                                                  \\ \hline
\end{tabular}
}
	\vspace{5pt}
	\caption{Appealing rate of black-box approximation under different accuracy requirements on CIFAR-10}
	\label{tb:black}
 	\vspace{-10pt}
\end{table*}

\subsection{Effectiveness of AppealNet}

In this subsection, we demonstrate the effectiveness of AppealNet in terms of the tradeoff between edge/cloud collaboration system accuracy (defined in Eq.~\eqref{eq:overall-acc}) and energy-saving (e.g. skipping rate defined in Eq.~\eqref{eq:sr} or computational cost defined in Eq.~\eqref{eq:overall-cost}).



First, we show the system accuracy under different skipping rate in Fig.~\ref{fig:acc-four}\footnote{Here, we only show our results when MobileNet is used as the small edge network. We see similar results with EfficientNet and ShuffleNet.}. As can be observed from this figure, the accuracy of AppealNet (the red line) is above the baseline methods in most cases across all three datasets. Such results are in line with our expectation as AppealNet can  differentiate `easy' and `difficult' inputs much better than baseline solutions. Take Fig.~\ref{fig:acc-four} (b) as an example, when the skipping rate is 95\% (\textit{i.e.}, 5\% of inputs are offloaded to the complex network), the system accuracy is 94.0\% with AppealNet while baseline solutions can only achieve 92.95\%. 
We can also see that the relative advantage of AppealNet over baseline solutions (the margin between the red line and other lines) is usually higher with increasing skipping rate (before reaching 100\%). This is because, the difference in such differentiation capabilities plays a more important role when the skipping rate is higher. 

One might expect that the overall accuracy of the edge/cloud system would be lower than a stand-alone large network since the small network is of low capacity. A seemingly surprising observation from Fig.~\ref{fig:acc-four} is that the edge/cloud system design could achieve accuracy boosting. Except for Tiny-ImageNet, the overall accuracy of the edge/cloud system can be above the accuracy of single ResNet-101 (dotted line) in certain skipping rate ranges. This is because, even though the small network has lower accuracy for the entire dataset, it is co-optimized with the big network to minimize the overall expected loss (as shown in Eq.~\eqref{eq:final-re}), and it is likely to correctly predict a portion of inputs that cannot be correctly classified by the big network. 
Compared to other confidence-based approaches, for the more difficult CIFAR-100 dataset, only AppealNet can generate higher accuracy than the standalone large network, as can be seen in Figure~\ref{fig:acc-four}(c). 
As for Tiny-ImageNet, the accuracy gap between the large and the small networks are too large (more than eight percent) to enable accuracy boosting when the skipping rate is in the range of $[70, 100]$. Under such circumstances, we could still benefit from AppealNet that provides a better tradeoff between prediction accuracy and energy savings when compared to other baseline solutions.

As we observe the score margin baseline performs slightly better than the other two baselines in most cases, in later results, we only compare AppealNet with it.


Table~\ref{tb:baseline} presents the comparison between baseline methods and AppealNet by calculating the overall computational cost (defined in Eq.~\eqref{eq:overall-cost}), measured in MFLOPs, under different accuracy improvement (defined in Eq.~\eqref{eq:acc-improvement}). The left part of the table present the accuracy and computational cost of the original ResNet-101, MobileNet and AppealNet, respectively. 
The key difference between AppealNet and MobileNet is that AppealNet has an extra predictor head and is optimized with a joint training scheme (See Section~\ref{subsec:jtm}), and the accuracy of the two-head AppealNet is close to the accuracy of MobileNet (See the second column in Table~\ref{tb:baseline}).


We now turn to the right part of Table~\ref{tb:baseline}, which shows the system computational cost with threshold $\delta$ tuned to match the target system accuracy. We evaluate four settings where the accuracy improvement of edge/cloud system over the standalone small edge DNN are set as \{50.0\%, 75.0\%, 90.0\%, 95.0\%\}, respectively. 
The relative cost savings of AppealNet over score margin baseline at different accuracy degradation settings are listed in the table. 
We can see that AppealNet outperforms in all these cases by a large margin. That is, we can further reduce the computation cost of edge/cloud collaborative system on GTSRB, CIFAR-10, CIFAR-100 and Tiny-ImageNet up to 35.77\%, 46.56\%, 43.60\% and 28.0\%, respectively.


\subsection{Black-Box Approximation}
\label{subsec:bl}
We further investigate the effectiveness of AppealNet under black-box settings, where the complex network is regarded as an oracle function without revealing its information. Since the oracle function always predicts correct results, the improvement of the overall accuracy comes from appealing those inputs that cannot be correctly classified by the small DNN to the cloud. As we cannot explicitly evaluate the computational cost of the whole system, the appealing rate is instead reported under accuracy improvement $AccI$ $\in$ \{50.0\%, 75.0\%, 90.0\%, 95.0\%\}. A lower appealing rate means less computations on the cloud and less communication cost, thereby saving more energy. As shown in Table~\ref{tb:black}, three efficient DNN models and their AppealNet counterparts are constructed and evaluated. The results in the right part of Table~\ref{tb:black} show the advantage of AppealNet over the baseline. For example, the relative saving of AppealNet on CIFAR-10 dataset under the black-box setting is 44.48\% at 50\% $AccI$ with EfficientNet as the small edge DNN architecture. 
\section{Conclusion}
\label{sec:conc}

In this paper, we propose AppealNet, an efficient and highly-accurate edge/cloud collaborative architecture for DNN-based classifiers. By employing a two-head architecture for the small edge network and jointly optimizing the approximator and the predictor in training, AppealNet can achieve trustworthy differentiation between easy and difficult inputs for the system, realizing up to more than 40\% energy savings compared to state-of-the-art solutions.


\section{Acknowledgement}

This work is supported in part by General Research Fund (GRF)
of Hong Kong Research Grants Council (RGC) under Grant No.
14205018 and No. 14205420, and in part by National  Natural  Science Foundation of China (NSFC) under Grant No. 61834006.



\begin{thebibliography}{10}
\providecommand{\url}[1]{#1}
\csname url@samestyle\endcsname
\providecommand{\newblock}{\relax}
\providecommand{\bibinfo}[2]{#2}
\providecommand{\BIBentrySTDinterwordspacing}{\spaceskip=0pt\relax}
\providecommand{\BIBentryALTinterwordstretchfactor}{4}
\providecommand{\BIBentryALTinterwordspacing}{\spaceskip=\fontdimen2\font plus
\BIBentryALTinterwordstretchfactor\fontdimen3\font minus
  \fontdimen4\font\relax}
\providecommand{\BIBforeignlanguage}[2]{{%
\expandafter\ifx\csname l@#1\endcsname\relax
\typeout{** WARNING: IEEEtran.bst: No hyphenation pattern has been}%
\typeout{** loaded for the language `#1'. Using the pattern for}%
\typeout{** the default language instead.}%
\else
\language=\csname l@#1\endcsname
\fi
#2}}
\providecommand{\BIBdecl}{\relax}
\BIBdecl

\bibitem{he2017channel}
Y.~He, X.~Zhang, and J.~Sun, ``Channel pruning for accelerating very deep
  neural networks,'' in \emph{Proceedings of the IEEE International Conference
  on Computer Vision}, 2017, pp. 1389--1397.

\bibitem{han2015deep}
S.~Han and et. al., ``Deep compression: Compressing deep neural network with
  pruning, trained quantization and huffman coding,'' in \emph{4th
  International Conference on Learning Representations (ICLR)}, 2016.

\bibitem{hinton2015distilling}
G.~Hinton, O.~Vinyals, and J.~Dean, ``Distilling the knowledge in a neural
  network,'' \emph{arXiv preprint arXiv:1503.02531}, 2015.

\bibitem{howard2017mobilenets}
A.~G. Howard, M.~Zhu, B.~Chen, D.~Kalenichenko, W.~Wang, T.~Weyand,
  M.~Andreetto, and H.~Adam, ``Mobilenets: Efficient convolutional neural
  networks for mobile vision applications,'' \emph{arXiv preprint
  arXiv:1704.04861}, 2017.

\bibitem{tann2016runtime}
H.~Tann, S.~Hashemi, and el. al., ``Runtime configurable deep neural networks
  for energy-accuracy trade-off,'' in \emph{2016 International Conference on
  Hardware/Software Codesign and System Synthesis (CODES+ ISSS)}.\hskip 1em
  plus 0.5em minus 0.4em\relax IEEE, 2016, pp. 1--10.

\bibitem{park2015big}
E.~Park and et. al., ``Big/little deep neural network for ultra low power
  inference,'' in \emph{2015 International Conference on Hardware/Software
  Codesign and System Synthesis (CODES+ISSS)}.\hskip 1em plus 0.5em minus
  0.4em\relax IEEE, 2015.

\bibitem{bolukbasi2017adaptive}
T.~Bolukbasi and et. al., ``Adaptive neural networks for efficient inference,''
  in \emph{Proceedings of the 34th International Conference on Machine
  Learning-Volume 70}.\hskip 1em plus 0.5em minus 0.4em\relax JMLR. org, 2017,
  pp. 527--536.

\bibitem{stamoulis2018design}
D.~Stamoulis, T.-W. Chin, A.~K. Prakash, H.~Fang, S.~Sajja, M.~Bognar, and
  D.~Marculescu, ``Designing adaptive neural networks for energy-constrained
  image classification,'' in \emph{Proceedings of the International Conference
  on Computer-Aided Design}, 2018, pp. 1--8.

\bibitem{hendrycks2016baseline}
D.~Hendrycks and K.~Gimpel, ``A baseline for detecting misclassified and
  out-of-distribution examples in neural networks,'' in \emph{5th International
  Conference on Learning Representations, {ICLR} 2017}, 2017.

\bibitem{guo2017calibration}
C.~Guo, G.~Pleiss, Y.~Sun, and K.~Q. Weinberger, ``On calibration of modern
  neural networks,'' in \emph{Proceedings of the 34th International Conference
  on Machine Learning-Volume 70}.\hskip 1em plus 0.5em minus 0.4em\relax JMLR.
  org, 2017.

\bibitem{lakshminarayanan2017simple}
B.~Lakshminarayanan, A.~Pritzel, and C.~Blundell, ``Simple and scalable
  predictive uncertainty estimation using deep ensembles,'' in \emph{Advances
  in neural information processing systems}, 2017, pp. 6402--6413.

\bibitem{wang2018skipnet}
X.~Wang, F.~Yu, and et. al., ``Skipnet: Learning dynamic routing in
  convolutional networks,'' in \emph{Proceedings of the European Conference on
  Computer Vision (ECCV)}, 2018, pp. 409--424.

\bibitem{wang2020dual}
Y.~Wang, J.~Shen, and et. al., ``Dual dynamic inference: Enabling more
  efficient, adaptive and controllable deep inference,'' \emph{IEEE Journal of
  Selected Topics in Signal Processing}, 2020.

\bibitem{panda2016conditional}
P.~Panda, A.~Sengupta, and K.~Roy, ``Conditional deep learning for
  energy-efficient and enhanced pattern recognition,'' in \emph{Design,
  Automation \& Test in Europe Conference \& Exhibition (DATE)}.\hskip 1em plus
  0.5em minus 0.4em\relax IEEE, 2016.

\bibitem{long2020conditionally}
Y.~Long, I.~Chakraborty, and K.~Roy, ``Conditionally deep hybrid neural
  networks across edge and cloud,'' \emph{arXiv:2005.10851}, 2020.

\bibitem{teerapittayanon2016branchynet}
S.~Teerapittayanon, B.~McDanel, and H.-T. Kung, ``Branchynet: Fast inference
  via early exiting from deep neural networks,'' in \emph{23rd International
  Conference on Pattern Recognition (ICPR)}.\hskip 1em plus 0.5em minus
  0.4em\relax IEEE, 2016.

\bibitem{liu2018dynamic}
L.~Liu and J.~Deng, ``Dynamic deep neural networks: Optimizing
  accuracy-efficiency trade-offs by selective execution,'' in
  \emph{Thirty-Second AAAI Conference on Artificial Intelligence}, 2018.

\bibitem{kaya2018shallow}
Y.~Kaya and et. al., ``Shallow-deep networks: Understanding and mitigating
  network overthinking,'' in \emph{Proceedings of the 36th International
  Conference on Machine Learning, {ICML}}.\hskip 1em plus 0.5em minus
  0.4em\relax {PMLR}, 2019.

\bibitem{yang2020resolution}
L.~Yang, Y.~Han, and et. al., ``Resolution adaptive networks for efficient
  inference,'' in \emph{Proceedings of the IEEE/CVF Conference on Computer
  Vision and Pattern Recognition}, 2020, pp. 2369--2378.

\bibitem{he2016deep}
K.~He, X.~Zhang, S.~Ren, and J.~Sun, ``Deep residual learning for image
  recognition,'' in \emph{Proceedings of the IEEE conference on computer vision
  and pattern recognition}, 2016, pp. 770--778.

\bibitem{nan2017adaptive}
F.~Nan and V.~Saligrama, ``Adaptive classification for prediction under a
  budget,'' in \emph{Proceedings of the 31st International Conference on Neural
  Information Processing Systems}, 2017, pp. 4730--4740.

\bibitem{kuhn2014nonlinear}
H.~W. Kuhn and A.~W. Tucker, ``Nonlinear programming,'' in \emph{Traces and
  emergence of nonlinear programming}.\hskip 1em plus 0.5em minus 0.4em\relax
  Springer, 2014, pp. 247--258.

\bibitem{TanL19}
M.~Tan and Q.~V. Le, ``Efficientnet: Rethinking model scaling for convolutional
  neural networks,'' in \emph{Proceedings of the 36th International Conference
  on Machine Learning, {ICML} 2019}, 2019.

\bibitem{ZhangZLS18}
X.~Zhang, X.~Zhou, M.~Lin, and J.~Sun, ``Shufflenet: An extremely efficient
  convolutional neural network for mobile devices,'' in \emph{{IEEE} Conference
  on Computer Vision and Pattern Recognition, {CVPR}}, 2018.

\bibitem{stallkamp2012man}
J.~Stallkamp, M.~Schlipsing, J.~Salmen, and C.~Igel, ``Man vs. computer:
  Benchmarking machine learning algorithms for traffic sign recognition,''
  \emph{Neural networks}, vol.~32, pp. 323--332, 2012.

\bibitem{krizhevsky2009learning}
A.~Krizhevsky, G.~Hinton \emph{et~al.}, ``Learning multiple layers of features
  from tiny images,'' Citeseer, Tech. Rep., 2009.

\bibitem{deng2009imagenet}
J.~Deng, W.~Dong, R.~Socher, L.-J. Li, K.~Li, and L.~Fei-Fei, ``Imagenet: A
  large-scale hierarchical image database,'' in \emph{2009 IEEE conference on
  computer vision and pattern recognition}.\hskip 1em plus 0.5em minus
  0.4em\relax IEEE, 2009, pp. 248--255.

\end{thebibliography}

\end{document}